\documentclass{article}
\usepackage{ijcai21}

\usepackage{times}
\usepackage{soul}
\usepackage{url}
\usepackage[hidelinks]{hyperref}
\usepackage[utf8]{inputenc}
\usepackage[small]{caption}
\usepackage{graphicx}
\usepackage{amsmath}
\usepackage{amsthm}
\usepackage{booktabs}
\usepackage{algorithm}
\usepackage{algorithmic}
\usepackage{amsthm,amssymb,mathrsfs}
\usepackage{multirow}
\usepackage{color}
\usepackage{bm}

\newcommand{\todo}[1]{{\color{red}#1}}

\title{
Unsupervised Knowledge Graph Alignment by Probabilistic Reasoning and Semantic Embedding
}

\author{
Zhiyuan Qi$^1$\and
Ziheng Zhang$^1$\and
Jiaoyan Chen$^2$\and
Xi Chen$^{1,3}$\footnote{Corresponding author (jasonxchen@tencent.com)}\and \\
Yuejia Xiang$^{1,3}$\and
Ningyu Zhang$^4$\And
Yefeng Zheng$^1$ \\
\affiliations
$^1$Tencent Jarvis Lab, Shenzhen, China\\
$^2$Department of Computer Science, University of Oxford, UK\\
$^3$Platform and Content Group, Tencent, Shenzhen, China\\
$^4$Zhejiang University, Hangzhou, China\\
}

\begin{document}

\maketitle

\begin{abstract}
Knowledge Graph (KG) alignment is to discover the mappings (i.e., equivalent entities, relations, and others) between two KGs.
The existing methods can be divided into the embedding-based models and the conventional reasoning and lexical matching based systems.
The former compute the similarity of entities via their cross-KG embeddings, but they usually rely on an ideal supervised learning setting for good performance and lack appropriate reasoning to avoid logically wrong mappings; while the latter address the reasoning issue but are poor at utilizing the KG graph structures and the entity contexts.
In this study, we aim at combining the above two solutions and thus propose an iterative framework named PRASE that is based on \textbf{p}robabilistic \textbf{r}easoning \textbf{a}nd \textbf{s}emantic \textbf{e}mbedding.
It learns the KG embeddings via entity mappings from a probabilistic reasoning system named PARIS, and feeds the resultant entity mappings and embeddings back into PARIS for augmentation.
The PRASE framework is compatible with different embedding-based models, and our experiments on multiple datasets have demonstrated its state-of-the-art performance.

\end{abstract}

\section{Introduction}
A knowledge graph (KG) organizes entities, attributes, relations, and other information in a structured format \cite{AEM2020ARXIV}. One single KG is often incomplete while different KGs can complement each other to form a larger and more comprehensive KG via alignment, i.e., discovering equivalent entities, relations, and others across two KGs.
%(a.k.a. mappings).
Due to wide KG applications, KG alignment, especially entity alignment, has attracted massive attention.

% With the development of machine learning, KG embeddings that encode entities, relations, and others into a low-dimensional vector space with their semantics preserved have become a powerful tool to manipulate and exploit KGs~\cite{wang2017knowledge}.
KG embeddings have become a powerful tool to exploit KGs by encoding entities, relations, and others into a low-dimensional vector space~\cite{wang2017knowledge}.
Many embedding-based models have been proposed for entity alignment~\cite{ZQW2020VLDB}, and usually comply with the following paradigm. They first embed the to-be-aligned KGs into one vector space and then discover the mappings by calculating the vector distance or similarity.

Although the embedding-based models have achieved encouraging results, they are still limited in some aspects especially in industrial deployment. These models usually require a number of known mappings (i.e., alignment seeds) for training.
However, seed annotation requires massive manual work, which may not be available in practice.
An industrial evaluation study has shown the number and the sampling distribution of alignment seeds can dramatically influence the alignment performance~\cite{ZJX2020COLING}.
The embedding-based models emphasize establishing expressive embeddings to capture entity features and then independently predict each mapping, ignoring the holistic analysis and logical consistency, which often leads to some false mappings.

In contrast, conventional KG alignment systems exploit various more traditional techniques such as logical reasoning and lexical matching. 
For example, the classic system LogMap~\cite{EB2011ISWC} iteratively discovers mappings by lexical and graph matching, and repairs mappings by logical reasoning. PARIS~\cite{FSP2012VLDB} is another representative system that utilizes probabilistic reasoning and lexical matching. 
Specifically, after getting initial mappings by matching with attributes such as names, PARIS expands the entity and relation mappings in each iteration by inferring the entity and relation equivalence with probabilistic reasoning. 
As no training is needed, these systems never rely on any alignment seeds, and are quite scalable and efficient. It is worth noting that 
PARIS and LogMap often outperform those embedding-based models according to the recent studies \cite{ZQW2020VLDB,ZJX2020COLING}. On the other hand, these conventional systems use traditional lexical and graph matching techniques that are weak at exploiting and utilizing the graph structure and other contextual information.

In light of the complementarity between the embedding-based models and conventional systems, we propose to construct a unified framework that absorbs the advantages of both. The main challenge is to find the effective ways to make two completely different models work together. In this work, an unsupervised iterative framework named PRASE is proposed, which is composed of a Probabilistic Reasoning (PR) module and a Semantic Embedding (SE) module. Specifically, the PR module initializes the mappings and infers logically consistent mappings with the entity embeddings from the SE module,
while the SE module emphasizes learning high-quality cross-KG embeddings that encode the graph structures and the entity contexts. Note that the SE module is compatible with all kinds of embedding-based alignment models; while the PR module is currently developed based on the conventional system PARIS, but it can be extended to other reasoning-based systems such as LogMap.

% The main contributions are summarized as follows.
The contributions of this paper are threefold.
First, an unsupervised KG alignment framework termed PRASE is proposed, which integrates probabilistic reasoning and semantic embedding using an iterative algorithm. 
To the best of our knowledge, this is the first to combine traditional reasoning techniques and state-of-the-art embedding techniques for KG alignment. Second, the PRASE framework has been implemented with PARIS and multiple different embedding-based models. Third, the PRASE framework has been evaluated on five widely used datasets and one industry dataset.
The results show the state-of-the-art performance of PRASE. 
On average, the F1-score of PRASE is $28.6\%$ higher than the best embedding-based model and $5.96\%$ higher than the best conventional system. 
Different settings of PRASE, such as the feedback from the SE module to the PR module and the iteration number, have also been studied.

\section{Preliminaries}
\label{background}
This section introduces the relevant background and the related work. The problem formulation is first given, and then the conventional system PARIS and the embedding-based models are briefly introduced.

% The section introduces the relevant background, including the problem formulation, the conventional system PARIS, and some embedding-based models.

\subsection{Problem Formulation}
Let $E$, $R$, $A$, and $V$ be the sets of entities, relations, attributes, and attribute values, respectively. A KG can be formulated as 
$G=(E,R,A,V,T^{\text{R}},T^{\text{A}})$,
where $T^{\text{R}}$ denotes the relation triples and $T^{\text{A}}$ represents the attribute triples. Specifically, $T^{\text{R}}$ and $T^{\text{A}}$ are formalized as
\begin{equation*}
    \small
	\begin{aligned}
		&T^{\text{R}} = \{(h, r, t)|h, t\in E, r\in R\},\\
		&T^{\text{A}} = \{(e, a, v)|e\in E, a\in A, v\in V\}.\\
	\end{aligned}
\end{equation*}

Given two KGs $G$ and $G'$, the problem of \textbf{Entity Alignment} is to discover the set of equivalent entity pairs (mappings) across $G$ and $G'$, denoted as
$$\mathcal{Y}=\{(e,e')|e\equiv e', e\in E, e'\in E'\},$$
where the equivalence $\equiv$ indicates that two entities refer to the same real-world object.

\subsection{PARIS}
Since the attribute triples are processed in a very similar way as the relation triples in PARIS,
for convenience, we define $E^{+}=E\cup V$, $R^{+}=R\cup A$, and $T^{+}=T^{\text{R}}\cup T^{\text{A}}$. In order to derive mappings, PARIS measures the functionality and inverse functionality of each relation, i.e.,
\begin{equation}
\label{PARIS_init}
\small
\begin{split}
	\text{F}(r) := \dfrac{|\{h|(h,r,t)\in T^{+}\}|}{|\{(h,t)|(h,r,t)\in T^{+}\}|},~r\in R^{+}, \\
	\text{F}^{-1}(r) := \dfrac{|\{t|(h,r,t)\in T^{+}\}|}{|\{(h,t)|(h,r,t)\in T^{+}\}|},~r\in R^{+},
\end{split}
\end{equation} 
where  $|\cdot|$ denotes the set cardinality.
The relation functionality and inverse functionality are used to determine the uniqueness of the head entity and tail entity, respectively.
Take the relation \textit{founder} as an example,
if the relation functionality is equal to one (i.e., $\text{F}(\textit{founder})=1$), it means that, given an organization, its founder can be uniquely determined.
Note that the functionality and the inverse functionality of the relations are invariant for a given KG and can be computed in advance.

PARIS alternately computes the entity mappings and the subsumption relationships between relations. In computing the entity mappings, 
the probability of equivalence between two entities $h$ and $h'$, denoted by $\text{P}(h\equiv h')$, is estimated as\footnote{Since PARIS augments the to-be-aligned KGs with the inverse triples, the formula actually accounts for both head and tail entities.}
\begin{equation}
	\label{PARIS_P_e}
	\small
	\begin{aligned}
		%\text{P}(h\equiv h') := 
		1 - \hspace{-2.3em}&\prod_{(h,r,t)\in T^{+},(h',r',t')\in T'^{+}}\hspace{-3.7em}\left(1 - \text{P}(r'\subseteq r)\text{F}^{-1}(r)\text{P}(t\equiv t')\right)\\&\hspace{5em}\times\left(1 - \text{P}(r\subseteq r')\text{F}^{-1}(r')\text{P}(t\equiv t')\right),
	\end{aligned}
\end{equation}
where $\text{P}(r\subseteq r')$ represents the probability that $r$ is a sub-relation of $r'$. 
$\text{P}(r\subseteq r')$ is computed as
\begin{equation}
    \small
	\begin{aligned}
		\label{PARIS_P_r}
		\dfrac{\sum_{h,t}\left(1 - \prod_{(h',r',t')\in T'^{+}}\left(1 - \text{P}(h\equiv h')\text{P}(t\equiv t')\right)\right)}{\sum_{h,t}\left(1 - \prod_{h',t'\in E'^{+}}\left(1 - \text{P}(h\equiv h')\text{P}(t\equiv t')\right)\right)},
	\end{aligned}
\end{equation}
where $(h,r,t)\in T^{+}.$ Similarly, $\text{P}(r'\subseteq r)$ can also be computed.
Note that the estimation of $\text{P}(h\equiv h')$ relies on the subsumption relationships between relations, i.e., $\text{P}(r\subseteq r')$ and $\text{P}(r'\subseteq r)$, and vice versa. 
Therefore, PARIS adopts an iterative strategy for optimization. In the initialization phase, $\text{P}(r\subseteq r')$ is set to a small value, e.g., $0.1$; $\text{P}(v\equiv v')~(v\in V\subseteq E^{+}~\text{and}~v'\in V'\subseteq E'^{+})$ is set to $1$ if $v$ and $v'$ are identical literals, and to $0$ otherwise.
Although this initialization method is simple, it has been shown quite effective. Other advanced methods, e.g., using the edit distance between text literals to score the equivalence of attribute values, can also be adopted.
In each iteration, the equivalence probabilities of entities are computed based on Eq.~\eqref{PARIS_P_e}, and then the probabilities for the subsumption relationships between relations are computed based on Eq.~\eqref{PARIS_P_r}. 
The system self-iterates multiple times until convergence.
Finally, the PARIS system outputs the entity mappings, denoted by $\tilde{\mathcal{Y}}^{\text{P}}$, along with their probabilities (equivalence degrees), denoted by $\text{P}^{\text{o}}(e\equiv e')~\text{with}~(e,e')\in\tilde{\mathcal{Y}}^{\text{P}}$ and superscript $^{\text{o}}$ indicating output.
Please see \cite{FSP2012VLDB} for more details.

\subsection{Embedding-based KG Alignment}
Embedding-based KG alignment models usually work in the following two steps. 
First, the embeddings of KG components are learned based on some translational models (e.g., TransE~\cite{ANA2013NIPS}), graph neural networks~\cite{TM2017ICLR} or other KG embedding algorithms~\cite{LZW2019ICML}. 
Entities of different KGs are embedded in the same vector space through strategies including parameter sharing, parameter swapping, embedding transformation, and embedding calibration.
Then, entity mappings are predicted based on the similarity measure of the entity embeddings.

Take the typical embedding-based model MTransE~\cite{MYM2017IJCAI} as an example. First, MTransE adopts TransE to learn embeddings by minimizing the following loss:
\begin{equation*}
    \small
	%\min~
	\sum_{(h,r,t)\in T^{\text{R}}}{||\mathbf{h}+\mathbf{r}-\mathbf{t}||} + \sum_{(h',r',t')\in T'^{\text{R}}}{||\mathbf{h}'+\mathbf{r}'-\mathbf{t}'||},
\end{equation*}
where $||\cdot||$ denotes the Euclidean norm operation; $\mathbf{h}$, $\mathbf{r}$, and $\mathbf{t}$ denote the $m$-dimensional embeddings of $h$, $r$, and $t$, respectively. 
To ensure the entities are embedded in the same vector space, an embedding transformation strategy is adopted. Let $\mathcal{S}=\{(e,e')|e\in E,e'\in E'\}$ be the alignment seeds, it minimizes the loss given by
$
    \small
	\sum_{(e,e')\in \mathcal{S}}~||M\mathbf{e}-\mathbf{e}'||,
$
where $\mathbf{e}$ and $\mathbf{e}'$ denote the entity embeddings, and $M\in\mathbb{R}^{m\times m}$ is a transformation matrix. After learning $M$, entity embeddings of $G$ are transformed into the entity vector space of $G'$, i.e., $\mathbf{e} :=M\mathbf{e}$. 
Finally, entity mappings, denoted by $\tilde{\mathcal{Y}}^{\text{E}}$, and their corresponding similarity scores in $\left[0, 1\right]$, denoted by $\text{S}(e\equiv e')~\text{with}~(e,e')\in\tilde{\mathcal{Y}}^{\text{E}}$, can be obtained by performing nearest neighbor search in the embedding space.

There are some other embedding-based models. For example,
% IPTransE~\cite{HRZ2017IJCAI} uses PTransE~\cite{YZH2015EMNLP} as the translational model.
GCN-Align embeds the KGs using graph convolutional networks~\cite{ZQX2018EMNLP}.
BootEA adopts a bootstrapping strategy with an alignment editing method to reduce error accumulation, so as to overcome the lack of training data~\cite{ZWQ2018IJCAI}. 
MultiKE embeds the entities with their names, relations, and attributes considered via multi-view learning~\cite{QZW2019IJCAI}.
Please see \cite{ZQW2020VLDB,ZJX2020COLING} and for more comprehensive reviews.

\section{Framework}
\label{framework}
In this section, the overview of the PRASE framework is given at first, and then its probabilistic reasoning module and semantic embedding module are introduced with details.
% In this section, we present the overview of the PRASE framework and then introduce its probabilistic reasoning and semantic embedding modules.

\subsection{PRASE Overview}
\begin{figure}[t!]
	\centering
	\includegraphics[width=1.0\linewidth]{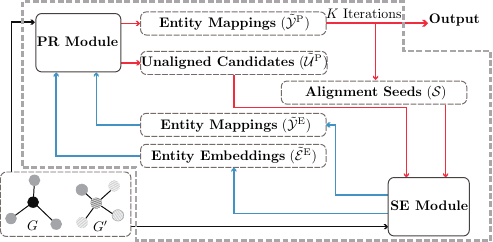}
	\caption{Overview of the PRASE framework.}
	\label{fig:PRASE}
\end{figure}

Figure \ref{fig:PRASE} shows the workflow of the PRASE framework that includes the
Probabilistic Reasoning (PR) module and the Semantic Embedding (SE) module. 
First, the PARIS-based PR module is performed on the input KGs to discover entity mappings $\tilde{\mathcal{Y}}^{\text{P}}$ with their probabilities $\text{P}^{\text{o}}(e\equiv e')$. 
Then, the highly confident entity mappings are selected as the alignment seeds $\mathcal{S}$, and the SE module is trained subsequently based on these seeds. After training, the SE module predicts mappings on the entities that have not been aligned by the PR module (denoted by $\tilde{\mathcal{U}}^{\text{P}}$). The resultant entity mappings $\tilde{\mathcal{Y}}^{\text{E}}$ with the similarity scores $\text{S}(e\equiv e')$ and the entity embeddings (denoted by $\tilde{\mathcal{E}}^{\text{E}}=\{\mathbf{e}|e\in E\}\cup\{\mathbf{e'}|e'\in E'\}$) are then fed back to the PR module. The above process can be iterated $K$ times, and the PR module finally outputs the entity mappings. The iterative algorithm is also shown in Algorithm \ref{alg:PRASE}. 

\subsection{Probabilistic Reasoning Module}
The PR module is constructed based on the PARIS system with the augmentation from KG embeddings.
Let $k$ be the iteration index of PRASE. In the initial iteration ($k=0$), the PR module first computes relation functionality and inverse functionality according to Eq.~(\ref{PARIS_init}), and then alternatively computes mappings following Eqs.~(\ref{PARIS_P_e}) and (\ref{PARIS_P_r}). 
In the subsequent iterations ($k=1,\cdots,K$), the SE and the PR modules are alternately performed. In the $k^{\text{th}}$ iteration ($k > 0$), the PR module is initialized based on its previous output and the output of the SE module. Specifically, the probabilities of sub-relationships are set to the values of the $(k-1)^{\text{th}}$ iteration, i.e., $\text{P}^{\text{i}}_{k}(r\subseteq r')=\text{P}^{\text{o}}_{k-1}(r\subseteq r')$ with superscripts $^{\text{i}}$ and $^{\text{o}}$ indicating the meanings of input and output, respectively, and the probabilities of entity mappings are initialized as
\begin{equation}
	\label{PRASE_init}
	\text{P}_{k}^{\text{i}}(e\equiv e') = \left\{\begin{array}{l}
		\alpha_{1}\text{P}_{k-1}^{\text{o}}(e\equiv e')~~~\hfill~\text{if}~(e,e')\in\tilde{\mathcal{Y}}^{\text{P}}_{k-1};
		\\\alpha_{2}\text{S}_{k}(e\equiv e')~\text{else if}~(e,e')\in\tilde{\mathcal{Y}}^{\text{E}}_{k}~\\\hfill\text{and}~\text{S}_{k}(e\equiv e')>\delta_1;
		\\  0\hfill\text{otherwise},
	\end{array}\right.
\end{equation} 
where $\text{P}^{\text{i}}$ and $\text{P}^{\text{o}}$ represent the input and the output mapping probabilities of the PR module, respectively; $\alpha_{1},\alpha_{2}\in\left(0,1\right]$ are two hyperparameters; $\delta_1 \in\left[0,1\right)$ is a threshold value; subscripts $_k$ and $_{k-1}$ indicate the variables in the $k^{\text{th}}$ and the $(k-1)^{\text{th}}$ iterations, respectively.
\iffalse{$\text{P}_{\text{e}}(e,e')$}\fi
% defined in Eq (5)?
\iffalse{What are $\text{P}_{\text{e}}(e,e')$ and $\text{P}_{k-1}^{\text{o}}(e,e')$? What's the annotation for the output mappings from the embedding?}\fi 
This customized initialization method for PARIS directly exploits the output of the SE module, through which the PR module could absorb the benefits captured by KG embeddings and further expand the mappings by reasoning. 
Note that the PR module is robust and can correct those unreliable entity mappings from the SE module, since it is based on probabilistic reasoning. 

In addition to directly utilizing the output from the SE module, we also seek to fully exploit the semantic and structural information contained in the embeddings during the self-iterations of the PR module. 
The probability of the equivalence between two entities is updated according to a modified estimation, which is given by
\begin{equation}
    \small
	\label{PRASE_P_e}
	\begin{aligned}
		\text{P}(e\equiv e'):=&(1-\beta)\cdot\text{sim}(\mathbf{e},\mathbf{e}') + \beta\Big(
		1 - \\\hspace{-1em}\prod_{(e,r,t)\in T^{+},(e',r',t')\in T'^{+}}\hspace{-3em}&{\left(1 - \text{P}(r'\subseteq r)\text{F}^{-1}(r)\text{P}(t\equiv t')\right)}\\\times&\left(1 - \text{P}(r\subseteq r')\text{F}^{-1}(r')\text{P}(t\equiv t')\right)\Big),
	\end{aligned}
\end{equation}
where $\mathbf{e}$ and $\mathbf{e}'$ denote the embeddings of entities $e$ and $e'$, respectively; $\text{sim}(\cdot,\cdot)$ is a similarity function; $\beta\in(0,1)$ is a trade-off hyperparameter balancing the embedding similarity and the probability estimated by Eq.~(\ref{PARIS_P_e}).
Eq.~(\ref{PRASE_P_e}) replaces Eq.~(\ref{PARIS_P_e}) in the original PARIS. It complements the PR module with the deep structural information learned from the SE model, and it can help the PR module estimate the probability of entity mappings more accurately. The function $\text{sim}(\cdot,\cdot)$ measures the similarity between two embeddings, of which the range should be $\left[0,1\right]$. A simple choice of $\text{sim}(\cdot,\cdot)$ is cosine similarity, i.e., $\text{sim}(\mathbf{e},\mathbf{e}')=\left(\mathbf{e}\cdot\mathbf{e}'\right)/\left(||\mathbf{e}||\cdot||\mathbf{e}'||\right)$. 

After several self-iterations, the PR module converges and outputs a new round of alignment results $\tilde{\mathcal{Y}}_{k}^{\text{P}}$ as well as the unaligned candidates $\tilde{\mathcal{U}}_{k}^{\text{P}}$. Specifically, $\tilde{\mathcal{U}}_{k}^{\text{P}}=\{e|e\in E,\forall e'\in E',(e,e')\notin\tilde{\mathcal{Y}}_{k}^{\text{P}}\}\cup\{e'|e'\in E',\forall e\in E,(e,e')\notin\tilde{\mathcal{Y}}_{k}^{\text{P}}\}$ is a set of unaligned entities,
which is later used as the test data for the SE module to compute the similarity score $\text{S}_{k+1}(e\equiv e')$. 
In the last iteration ($k=K$), the final output entity mappings of the PRASE framework are given by the PR module, denoted as $\tilde{\mathcal{Y}}_{\text{f}}=\{(e,e')|(e,e')\in\tilde{\mathcal{Y}}_{K}^{\text{P}},\text{P}_{K}^{\text{o}}(e\equiv e')>\delta_{\text{f}}\},$ where $\delta_{\text{f}}\in\left[0,1\right)$ is a threshold value.

\begin{algorithm}[tb]
	\caption{PARIS-based PRASE Implementation}
	\label{alg:PRASE}
	\textbf{Input}: two KGs $G$ and $G'$\\
	\textbf{Parameter}: iteration number $K$, hyperparameters $\alpha_{1}$, $\alpha_{2}$, and $\beta$, thresholds $\delta_{1}$, $\delta_{2}$, and $\delta_{\text{f}}$, similarity function $\text{sim}(\cdot,\cdot)$
	
	\begin{algorithmic}[1] 
		\STATE Initialize the PR module using Eq. (\ref{PARIS_init});
		\STATE Perform the PR module using Eqs. (\ref{PARIS_P_e}) and (\ref{PARIS_P_r});
		\STATE Generate $\tilde{\mathcal{Y}}^{\text{P}}_{0}$ and $\tilde{\mathcal{U}}^{\text{P}}_{0}$;
		\WHILE{$k=1,\ldots,K$}
		\STATE Generate $\mathcal{S}_{k}$ based on $\tilde{\mathcal{Y}}^{\text{P}}_{k-1}$ and $\delta_{2}$;
		\STATE Train the SE module on $\mathcal{S}_{k}$; 
		\STATE Test the SE module on $\tilde{\mathcal{U}}^{\text{P}}_{k-1}$;
		\STATE Generate $\tilde{\mathcal{Y}}^{\text{E}}_{k}$ and  $\tilde{\mathcal{E}}^{\text{E}}_{k}$;
		\STATE Initialize the PR module using Eq. (\ref{PRASE_init});
		\STATE Perform the PR module using Eqs. (\ref{PRASE_P_e}) and (\ref{PARIS_P_r});
		\STATE Generate $\tilde{\mathcal{Y}}^{\text{P}}_{k}$ and $\tilde{\mathcal{U}}^{\text{P}}_{k}$;
		\ENDWHILE
		\STATE Generate $\tilde{\mathcal{Y}}_{\text{f}}$ based on $\tilde{\mathcal{Y}}^{\text{P}}_{K}$ and $\delta_{\text{f}}$;
	\end{algorithmic}
	\textbf{Output}: $\tilde{\mathcal{Y}}_{\text{f}}$
\end{algorithm}

\begin{table}[t]
	\centering
	\scriptsize
	\renewcommand\arraystretch{0.1}
	\setlength{\tabcolsep}{2mm}{
		\begin{tabular}{@{}c|c|c|c|c|c|c@{}}
			\toprule
			\multirow{2}{*}{Dataset}    & \multirow{2}{*}{KGs} & \multirow{2}{*}{\#Ents.} & \multicolumn{2}{c|}{Relation} & \multicolumn{2}{c}{Attribute} \\ \cmidrule(l){4-7} 
			&                      &                          & \#Rels.      & \#Triples      & \#Attrs.      & \#Triples      \\ \midrule
			\multirow{2}{*}{EN-FR-100K} & EN                   & 100,000                  & 379          & 649,902        & 364           & 503,922        \\ \cmidrule(l){2-7} 
			& FR                   & 100,000                  & 287          & 561,391        & 468           & 431,379        \\ \midrule
			\multirow{2}{*}{EN-DE-100K} & EN                   & 100,000                  & 323          & 622,588        & 326           & 560,247        \\ \cmidrule(l){2-7} 
			& DE                   & 100,000                  & 170          & 629,395        & 189           & 793,710        \\ \midrule
			\multirow{2}{*}{D-W-100K}   & DB                   & 100,000                  & 318          & 616,457        & 328           & 467,103        \\ \cmidrule(l){2-7} 
			& WD                   & 100,000                  & 239          & 588,203        & 760           & 878,219        \\ \midrule
			\multirow{2}{*}{D-Y-100K}   & DB                   & 100,000                  & 230          & 576,547        & 277           & 547,026        \\ \cmidrule(l){2-7} 
			& YG                   & 100,000                  & 31           & 865,265        & 36            & 855,161        \\ \midrule
			\multirow{2}{*}{D-W-15K}    & DB                   & 15,000                   & 167          & 73,983         & 175           & 66,813         \\ \cmidrule(l){2-7} 
			& WD                   & 15,000                   & 121          & 83,365         & 457           & 175,686        \\ \midrule
			\multirow{2}{*}{MED-BBK-9K} & MED                  & 9,162                    & 32           & 158,357        & 19            & 11,467         \\ \cmidrule(l){2-7} 
			& BBK                  & 9,162                    & 20           & 50,307         & 21            & 44,987         \\ \bottomrule
		\end{tabular}
	}
	\caption{Dataset statistics.}
	\label{tab:datasets}
\end{table}
\subsection{Semantic Embedding Module}
In the $k^{\text{th}}$ iteration, $\tilde{\mathcal{Y}}^{\text{P}}_{k-1}$ is refined to generate reliable alignment seeds $\mathcal{S}_{k}$ for training the SE module. A feasible method to obtain $\mathcal{S}_{k}$ is to set a threshold $\delta_{2}\in\left[0, 1\right)$, and $\mathcal{S}_{k}=\{(e,e')|(e,e')\in\tilde{\mathcal{Y}}^{\text{P}}_{k-1},\text{P}_{k-1}^{\text{o}}(e\equiv e')>\delta_{2}\}$. Although in most cases, there are still some incorrect entity mappings in $\mathcal{S}_{k}$, the abundant correct mappings in $\mathcal{S}_{k}$ can still provide useful information. 
As mentioned before, the SE module outputs \textit{(i)} entity mappings via a nearest neighbour search among $\tilde{\mathcal{U}}_{k-1}^{\text{P}}$,
and \textit{(ii)} entity embeddings $\tilde{\mathcal{E}}^{\text{E}}_{k}$. Since almost all embedding-based models can output entity mappings and embeddings, PRASE can choose almost any existing embedding-based model as the SE module. Algorithm \ref{alg:PRASE} shows the whole process of the PRASE framework.

\section{Evaluation}
\label{experiment}
This section presents the evaluation of PRASE, and the code is available at \url{https://github.com/qizhyuan/PRASE-Python}.

\subsection{Datasets}
In the experiments, the following datasets are used, and the statistics of these datasets are presented in Table \ref{tab:datasets}.

\noindent\textbf{OpenEA Datasets}: The OpenEA datasets\footnote{\url{https://github.com/nju-websoft/OpenEA}} are constructed based on DBpedia, YAGO, and Wikidata~\cite{ZQW2020VLDB}. 
We use all their large-scale datasets of the version ``V2'' that has more complex KG structures.
They include two cross-lingual datasets (i.e., EN-FR-100K-V2 and EN-DE-100K-V2) and two cross-KG datasets (i.e., D-W-100K-V2 and D-Y-100K-V2). We
also use a small dataset D-W-15K-V2, a relatively difficult dataset as reported by~\cite{ZQW2020VLDB}.
In the following, the annotation ``-V2'' is omitted.

\noindent\textbf{Industry Dataset}: MED-BBK-9K is an industry dataset\iffalse\footnote{\url{https://github.com/ZihengZZH/industry-eval-EA}}\fi proposed by \cite{ZJX2020COLING}, which is built from an authoritative medical KG and a KG extracted from Baidu Baike, a Chinese online encyclopedia.

\begin{table*}[]
	\scriptsize
	\centering
	\renewcommand{\arraystretch}{0.98}
	\setlength{\tabcolsep}{1.1mm}{
	\begin{tabular}{@{}ccccccccccccccccccc@{}}
		\toprule
		\multicolumn{1}{c|}{\multirow{2}{*}{Model}}           & \multicolumn{3}{c|}{EN-FR-100K}                                                                                                    & \multicolumn{3}{c|}{EN-DE-100K}                                                                               & \multicolumn{3}{c|}{D-W-100K}                                                                                                      & \multicolumn{3}{c|}{D-Y-100K}                                                                                                      & \multicolumn{3}{c|}{D-W-15K}                                                                                                       & \multicolumn{3}{c}{MED-BBK-9K}                                                                               \\ \cmidrule(l){2-19} 
		& \multicolumn{1}{|c}{P}                                  & R                                  & \multicolumn{1}{c|}{F1}                                 & P                                  & R                                  & \multicolumn{1}{c|}{F1}            & P                                  & R                                  & \multicolumn{1}{c|}{F1}                                 & P                                  & R                                  & \multicolumn{1}{c|}{F1}                                 & P                                  & R                                  & \multicolumn{1}{c|}{F1}                                 & P                                  & R                                  & F1                                 \\ \midrule
		\multicolumn{1}{c|}{MTransE}     & 0.090                              & 0.090                              & \multicolumn{1}{c|}{0.090}                              & 0.115                              & 0.115                              & \multicolumn{1}{c|}{0.115}         & 0.148                              & 0.148                              & \multicolumn{1}{c|}{0.148}                              & 0.100                              & 0.100                              & \multicolumn{1}{c|}{0.100}                              & 0.271                              & 0.271                              & \multicolumn{1}{c|}{0.271}                              & 0.002                              & 0.002                              & 0.002                              \\
		\multicolumn{1}{c|}{IPTransE}    & 0.234                              & 0.234                              & \multicolumn{1}{c|}{0.234}                              & 0.346                              & 0.346                              & \multicolumn{1}{c|}{0.346}         & 0.319                              & 0.319                              & \multicolumn{1}{c|}{0.319}                              & 0.456                              & 0.456                              & \multicolumn{1}{c|}{0.456}                              & 0.412                              & 0.412                              & \multicolumn{1}{c|}{0.412}                              & 0.054                              & 0.054                              & 0.054                              \\
		\multicolumn{1}{c|}{GCNAlign}    & 0.257                              & 0.257                              & \multicolumn{1}{c|}{0.257}                              & 0.375                              & 0.375                              & \multicolumn{1}{c|}{0.375}         & 0.353                              & 0.353                              & \multicolumn{1}{c|}{0.353}                              & 0.620                              & 0.620                              & \multicolumn{1}{c|}{0.620}                              & 0.506                              & 0.506                              & \multicolumn{1}{c|}{0.506}                              & 0.057                              & 0.057                              & 0.057                              \\
		\multicolumn{1}{c|}{BootEA}      & 0.640                              & 0.640                              & \multicolumn{1}{c|}{0.640}                              & 0.739                              & 0.739                              & \multicolumn{1}{c|}{0.739}         & 0.766                              & 0.766                              & \multicolumn{1}{c|}{0.766}                              & 0.886                              & 0.886                              & \multicolumn{1}{c|}{0.886}                              & 0.821                              & 0.821                              & \multicolumn{1}{c|}{0.821}                              & 0.307                              & 0.307                              & 0.307                              \\
		\multicolumn{1}{c|}{RSN4EA}       & 0.495                              & 0.495                              & \multicolumn{1}{c|}{0.495}                              & 0.639                              & 0.639                              & \multicolumn{1}{c|}{0.639}         & 0.634                              & 0.634                              & \multicolumn{1}{c|}{0.634}                              & 0.841                              & 0.841                              & \multicolumn{1}{c|}{0.841}                              & 0.723                              & 0.723                              & \multicolumn{1}{c|}{0.723}                              & 0.195                           & 0.195                              & 0.195                              \\
		\multicolumn{1}{c|}{IMUSE}       & 0.461                              & 0.461                              & \multicolumn{1}{c|}{0.461}                              & 0.457                              & 0.457                              & \multicolumn{1}{c|}{0.457}         & 0.431                              & 0.431                              & \multicolumn{1}{c|}{0.431}                              & 0.629                              & 0.629                              & \multicolumn{1}{c|}{0.629}                              & 0.581                              & 0.581                              & \multicolumn{1}{c|}{0.581}                              & 0.186                              & 0.186                              & 0.186                              \\
		\multicolumn{1}{c|}{MultiKE}     & 0.642                              & 0.642                              & \multicolumn{1}{c|}{0.642}                              & 0.661                              & 0.661                              & \multicolumn{1}{c|}{0.661}         & 0.319                              & 0.319                              & \multicolumn{1}{c|}{0.319}                              & 0.853                              & 0.853                              & \multicolumn{1}{c|}{0.853}                              & 0.495                              & 0.495                              & \multicolumn{1}{c|}{0.495}                              & 0.410                              & 0.410                              & 0.410                              \\
		\multicolumn{1}{c|}{RDGCN}       & 0.715                              & 0.715                              & \multicolumn{1}{c|}{0.715}                              & 0.766                              & 0.766                              & \multicolumn{1}{c|}{0.766}         & 0.421                              & 0.421                              & \multicolumn{1}{c|}{0.421}                              & 0.911                              & 0.911                              & \multicolumn{1}{c|}{0.911}                              & 0.623                              & 0.623                              & \multicolumn{1}{c|}{0.623}                              & 0.301                              & 0.301                              & 0.301                              \\ \midrule
		\multicolumn{1}{c|}{PARIS}       & \textbf{0.981}                              & 0.877                              & \multicolumn{1}{c|}{0.926}                              & \underline{0.988}                              & 0.912                              & \multicolumn{1}{c|}{0.948}         & \textbf{0.931}                              & 0.788                              & \multicolumn{1}{c|}{0.854}                              & 0.997                              & 0.970                              & \multicolumn{1}{c|}{0.983}                              & \textbf{0.950}                              & 0.850                              & \multicolumn{1}{c|}{0.897}                              & 0.779                              & 0.367                              & 0.499                              \\
		\multicolumn{1}{c|}{LogMap}      & 0.541                              & 0.709                              & \multicolumn{1}{c|}{0.614}                              & 0.729                              & 0.729                              & \multicolumn{1}{c|}{0.729}         & -                                  & -                                  & \multicolumn{1}{c|}{-}                                  & 0.954                              & 0.912                              & \multicolumn{1}{c|}{0.933}                              & -                                  & -                                  & \multicolumn{1}{c|}{-}                                  & \textbf{0.864}                                  & 0.441                                  & \underline{0.584}                                  \\ \midrule
		\multicolumn{1}{c|}{STR-Match}   & 0.754 & 0.720                              & \multicolumn{1}{c|}{0.737}                              & 0.790 & 0.760                              & \multicolumn{1}{c|}{0.775}         & 0.419 & 0.306                              & \multicolumn{1}{c|}{0.354}                              & \textbf{1.000} & \textbf{1.000} & \multicolumn{1}{c|}{\textbf{1.000}} & 0.606 & 0.419                              & \multicolumn{1}{c|}{0.495}                              & 0.545 & 0.495                              & 0.519                              \\
		\multicolumn{1}{c|}{EMB-Match} & 0.731 & 0.661 & \multicolumn{1}{c|}{0.694} & 0.747 & 0.694 & \multicolumn{1}{c|}{0.720} & 0.318 & 0.308 & \multicolumn{1}{c|}{0.313} & 0.866 & 0.838 & \multicolumn{1}{c|}{0.852} & 0.501 & 0.485 & \multicolumn{1}{c|}{0.493} & 0.504 & \underline{0.504} & 0.504 \\ \midrule
		\multicolumn{1}{c|}{PRASE-BootEA} & 0.977                              & \textbf{0.932} & \multicolumn{1}{c|}{\textbf{0.954}} & 0.983                              & \underline{0.948}                              & \multicolumn{1}{c|}{\underline{0.965}}         & \underline{0.927}                              & \textbf{0.855} & \multicolumn{1}{c|}{\textbf{0.890}} & 0.998                              & 0.989                              & \multicolumn{1}{c|}{0.993}                              & \underline{0.948}                              & \textbf{0.900} & \multicolumn{1}{c|}{\textbf{0.923}} & 0.687                              & 0.469                              & 0.557                              \\
		\multicolumn{1}{c|}{PRASE-MultiKE}                     & \underline{0.979}                              & \underline{0.930}                              & \multicolumn{1}{c|}{\underline{0.954}}                                                   & \textbf{0.988}                              & \textbf{0.955} & \multicolumn{1}{c|}{\textbf{0.972}} & 0.922                              & \underline{0.804}                              & \multicolumn{1}{c|}{\underline{0.859}}                                                   & \underline{0.998}                              & \underline{0.993}                              & \multicolumn{1}{c|}{\underline{0.996}}                                                   & 0.941                              & \underline{0.875}                              & \multicolumn{1}{c|}{\underline{0.907}}                                                   & \underline{0.837}                              & \textbf{0.619} & \textbf{0.711} \\ \bottomrule
	\end{tabular}
	}
	\caption{The overall results of the PRASE models in comparison with the baselines.}
	\label{tab:result}
\end{table*}
\begin{table*}[]
	\renewcommand\arraystretch{0.98}
	\centering
	\scriptsize
	\setlength{\tabcolsep}{1.1mm}{
		\begin{tabular}{@{}c|ccc|ccc|ccc|ccc|ccc|ccc@{}}
			\toprule
			\multirow{2}{*}{Model} & \multicolumn{3}{c|}{EN-FR-100K}                                                                              & \multicolumn{3}{c|}{EN-DE-100K}                                                                              & \multicolumn{3}{c|}{D-W-100K}                                                                                & \multicolumn{3}{c|}{D-Y-100K}                                                                                & \multicolumn{3}{c|}{D-W-15K}                                                                                 & \multicolumn{3}{c}{MED-BBK-9K}                                                                              \\ \cmidrule(l){2-19} 
			& P                                  & R                                  & F1                                 & P                                  & R                                  & F1                                 & P                                  & R                                  & F1                                 & P                                  & R                                  & F1                                 & P                                  & R                                  & F1                                 & P                                  & R                                  & F1                                 \\ \midrule
			PARIS                  & 0.981                              & 0.877                              & 0.926                              & 0.988                              & 0.912                              & 0.948                              & \textbf{0.931} & 0.788                              & 0.854                              & 0.997                              & 0.970                              & 0.983                              & \textbf{0.950} & 0.850                              & 0.897                              & 0.779                              & 0.367                              & 0.499                              \\ \midrule
			PRASE-BootEA-M          & 0.976                              & 0.912                              & 0.943                              & 0.983                              & 0.934                              & 0.958                              & 0.928                              & \underline{0.846}                              & \underline{0.885}                              & 0.997                              & 0.984                              & 0.991                              & 0.948                              & \underline{0.898}                              & \underline{0.923}                              & 0.692                              & 0.458                              & 0.552                              \\
			PRASE-BootEA-E          & \underline{0.982}                              & 0.912                              & 0.945                              & 0.987                              & 0.933                              & 0.960                              & 0.928                              & 0.821                              & 0.871                              & 0.997                              & 0.978                              & 0.988                              & 0.949                              & 0.876                              & 0.911                              & 0.761                              & 0.435                              & 0.554                              \\
			PRASE-BootEA            & 0.977                              & \textbf{0.932} & \textbf{0.954} & 0.983                              & \underline{0.948}                              & 0.965                              & 0.927                              & \textbf{0.855} & \textbf{0.890} & 0.998                              & 0.989                              & 0.993                              & 0.948                              & \textbf{0.900} & \textbf{0.923} & 0.687                              & 0.469                              & 0.557                              \\ \midrule
			PRASE-MultiKE-M         & 0.977                              & 0.914                              & 0.945                              & 0.987                              & 0.946                              & \underline{0.966}                              & 0.923                              & 0.800                              & 0.857                              & \underline{0.998}                              & \underline{0.990}                              & \underline{0.994}                              & 0.943                              & 0.877                              & 0.909                              & 0.825                              & \underline{0.593}                              & \underline{0.690}                              \\
			PRASE-MultiKE-E         & \textbf{0.984} & 0.903                              & 0.942                              & \textbf{0.989} & 0.930                              & 0.959                              & \underline{0.930}                              & 0.798                              & 0.859                              & 0.997                              & 0.978                              & 0.988                              & \underline{0.949}                              & 0.863                              & 0.904                              & \textbf{0.837} & 0.493                              & 0.621                              \\
			PRASE-MultiKE           & 0.979                              & \underline{0.930}                              & \underline{0.954}                              & \underline{0.988}                              & \textbf{0.955} & \textbf{0.972} & 0.922                              & 0.804                              & 0.859                              & \textbf{0.998} & \textbf{0.993} & \textbf{0.996} & 0.941                              & 0.875                              & 0.907                              & \underline{0.837}                              & \textbf{0.619} & \textbf{0.711} \\ \bottomrule
		\end{tabular}
	}
	\caption{The results of PRASE with different feedback settings from the SE module to the PR module.
	}
	\label{tab:vs}
\end{table*}
\subsection{Experimental Setting}
The original implementation of PARIS is in Java.\footnote{\url{http://webdam.inria.fr/paris/}}
We re-implemented PARIS in Python and updated it as the PR module such that it can easily work with the embedding-based models that are also implemented in Python.
We adopt twelve competitive KG alignment methods as the baselines.
They can be categorized into \textit{(i)} embedding-based models that include MTransE~\cite{MYM2017IJCAI}, IPTransE~\cite{HRZ2017IJCAI},  GCNAlign~\cite{ZQX2018EMNLP}, BootEA~\cite{ZWQ2018IJCAI}, RSN4EA~\cite{LZW2019ICML}, IMUSE~\cite{FZY2019DASFAA}, MultiKE~\cite{QZW2019IJCAI}, and RDGCN~\cite{YXY2019IJCAI}, \textit{(ii)} conventional systems including PARIS and LogMap, and 
\textit{(iii)} two simple matching models using either the edit distance (denoted by STR-Match) or the word embedding similarity (denoted by EMB-Match) between entity names.
We adopt the implementations of the embedding-based models from OpenEA~\cite{ZQW2020VLDB} with the same dataset division: 20\%, 10\%, and 70\% of the entity mappings for training, validation, and testing, respectively.
Except for embedding-based models, other models are performed in an unsupervised setting. 
STR-Match and EMB-Match compute the similarity between entity names with a threshold of $0.5$.
In the overall result analysis,  BootEA and MultiKE are used as the SE module and the resultant models are denoted by PRASE-BootEA and PRASE-MultiKE; while in the ablation study, more embedding-based models are evaluated.
We set $\alpha_{1}=\alpha_{2}=1$, $\beta=0.8$, $\delta_{1}=\delta_{2}=\delta_{\text{f}}=0.1$, and choose cosine similarity as $\text{sim}(\cdot)$.
%if they are not specified. 
Since a small value of $K$ is found to be sufficient for PRASE to demonstrate its effectiveness, we set $K=1$ in the experiments unless specified.

Our experiments are conducted on a workstation with an Intel Xeon E5 CPU and an NVIDIA Tesla M40 GPU.
The average time cost of our PARIS implementation on the four 100K datasets is 1697 seconds\footnote{The average running time by the Java implementation is 89 s.}, while the average time costs of BootEA and MultiKE are 24727 and 3198 seconds, respectively.
Therefore, the time cost of the PRASE framework is acceptable even with several iterations executed.
As a comprehensive metric, F1-score is used to evaluate different models in the following experiments with the corresponding precision (P) and recall (R) reported as supplementary metrics.
Since embedding-based models output a list of matching candidates for each entity, their precision, recall, and F1-score are actually equivalent to Hits@1.
The best performance is \textbf{bolded} and the second best is \underline{underlined} in our experiments. 

\begin{table*}[]
	\renewcommand\arraystretch{0.98}
	\centering
	\scriptsize
	\setlength{\tabcolsep}{1.1mm}{
		\begin{tabular}{@{}c|ccc|ccc|ccc|ccc|ccc|ccc@{}}
			\toprule
			\multirow{2}{*}{Model} & \multicolumn{3}{c|}{EN-FR-100K}                                                                              & \multicolumn{3}{c|}{EN-DE-100K}                                                                              & \multicolumn{3}{c|}{D-W-100K}                                                                                & \multicolumn{3}{c|}{D-Y-100K}                                                                                & \multicolumn{3}{c|}{D-W-15K}                                                                                 & \multicolumn{3}{c}{MED-BBK-9K}                                                                              \\ \cmidrule(l){2-19} 
			& P                                  & R                                  & F1                                 & P                                  & R                                  & F1                                 & P                                  & R                                  & F1                                 & P                                  & R                                  & F1                                 & P                                  & R                                  & F1                                 & P                                  & R                                  & F1                                 \\ \midrule
			PARIS                  & \textbf{0.981}                              & 0.877                              & 0.926                              & \underline{0.988}                              & 0.912                              & 0.948                              & \textbf{0.931} & 0.788                              & 0.854                              & 0.997                              & 0.970                              & 0.983                              & \textbf{0.950} & 0.850                              & 0.897                              & \underline{0.779}                              & 0.367                              & 0.499                              \\ \midrule
			PRASE-MTransE           & 0.970                              & 0.908                              & 0.938                              & 0.980                              & 0.933                              & 0.956                              & 0.914                              & 0.821                              & 0.865                              & 0.995                              & 0.984                              & 0.989                              & 0.944                              & \underline{0.886} & \underline{0.914} & 0.668                              & 0.381                              & 0.485                              \\
			PRASE-IPTransE          & 0.979                              & 0.918 & 0.947 & 0.985                              & 0.938 & 0.961 & 0.927                              & 0.825                              & 0.873                              & 0.997                              & 0.986                              & 0.992                              & 0.945                              & 0.879                              & 0.910                              & 0.650                              & 0.429 & 0.517 \\
			PRASE-GCNAlign          & \underline{0.981} & 0.900                              & 0.939                              & 0.986 & 0.924                              & 0.954                              & 0.927 & 0.803                              & 0.861                              & 0.997                              & 0.975                              & 0.986                              & \underline{0.950} & 0.863                              & 0.904                              & 0.676 & 0.418                              & 0.517                              \\
			PRASE-IMUSE             & 0.973                              & 0.912                              & 0.941                              & 0.982                              & 0.938                              & 0.959                              & 0.923                              & \underline{0.829} & \underline{0.873} & 0.997 & 0.987 & 0.992 & 0.945                              & 0.883                              & 0.913                              & 0.616                              & 0.422                              & 0.501                              \\
			PRASE-BootEA            & 0.977                              & \textbf{0.932} & \textbf{0.954} & 0.983                              & \underline{0.948}                              & \underline{0.965}                              & \underline{0.927}                              & \textbf{0.855} & \textbf{0.890} & \underline{0.998}                           & \underline{0.989}                              & \underline{0.993}                              & 0.948                              & \textbf{0.900} & \textbf{0.923} & 0.687                              & \underline{0.469}                              & \underline{0.557}                              \\
			PRASE-MultiKE           & 0.979                              & \underline{0.930}                              & \underline{0.954}                              & \textbf{0.988}                              & \textbf{0.955} & \textbf{0.972} & 0.922                              & 0.804                              & 0.859                              & \textbf{0.998} & \textbf{0.993} & \textbf{0.996} & 0.941                              & 0.875                              & 0.907                              & \textbf{0.837}                              & \textbf{0.619} & \textbf{0.711} \\
			\bottomrule
		\end{tabular}
	}
	\caption{The comparison results of the PRASE models using different embedding-based models.
	}
	\label{tab:PRASE_models}
\end{table*}
\subsection{Overall Results}
Table \ref{tab:result} presents the experimental results (here, we directly use the results from \cite{ZQW2020VLDB} for embedding-based models and LogMap on OpenEA datasets). 
The results show that the two PRASE models consistently outperform all the baselines on all the datasets except D-Y-100K in terms of recall and F1-score. STR-Match reaches full scores and outperforms all other models on D-Y-100K. However, the recall of STR-Match is significantly lower than the proposed models on the other five datasets. Actually, D-Y-100K is an easy dataset on which STR-Match achieves a perfect alignment, while the proposed models are almost perfect. \iffalse{In the following, we focus on analyzing results on the other five datasets.}\fi PRASE-BootEA performs best on EN-FR-100K, D-W-100K, and D-W-15K, while PRASE-MultiKE performs best on EN-DE-100K and MED-BBK-9K. Besides, the F1-score of PRASE-MultiKE reaches 0.711 on MED-BBK-9K, significantly surpassing PARIS by 0.212. Compared with PARIS, the precision of the proposed models is slightly decreased on EN-FR-100K, D-W-100K, and D-W-15K, but the recall significantly increases. It reflects that although the incorrect entity mappings predicted by the SE module can have a negative impact on the performance, the useful information provided by the SE module can still help the PR module find more potential alignments. Table \ref{tab:result} also shows that the unsupervised PRASE models significantly outperform all the supervised embedding-based models with an average improvement of 28.6\% in F1-score, which further confirms the effectiveness of the PRASE framework. 

\iffalse\todo{Directly report the time of PRASE (python)-BootEA and PRASE (python)-MultiKE. Report the time PARIS in java in the text.}\fi
\begin{table}[]
	\renewcommand\arraystretch{0.1}
	\scriptsize
	\centering
	\setlength{\tabcolsep}{0.8mm}{
		\begin{tabular}{@{}c|c|c|c|c|c|c@{}}
			\toprule
			\multirow{2}{*}{Dataset} & MTransE & IPTransE & GCNAlign & IMUSE & BootEA & MultiKE \\ \cmidrule(l){2-7} 
			& \multicolumn{6}{c}{Hits@1}                                                                                     \\ \midrule
			EN-FR-100K               & 0.281   & 0.439    & 0.328    & 0.382 & \textbf{0.629} & \underline{0.445}                              \\ \midrule
			EN-DE-100K               & 0.288   & 0.442    & 0.338    & 0.425 & \textbf{0.576} & \underline{0.575}                              \\ \midrule
			D-W-100K                 & 0.283   & 0.328    & 0.318    & \underline{0.345} & \textbf{0.522} & 0.117                              \\ \midrule
			D-Y-100K                 & 0.574   & 0.806    & 0.745    & 0.811 & \underline{0.858}                              & \textbf{0.906} \\ \midrule
			D-W-15K                  & 0.457   & 0.413    & 0.361    & \underline{0.481} & \textbf{0.614} & 0.364                              \\ \midrule
			MED-BBK-9K               & 0.009   & 0.139    & 0.070    & 0.120 & \underline{0.233}                              & \textbf{0.433} \\ \bottomrule
		\end{tabular}
	}
	\caption{The performance of different SE modules.
	}
	\label{tab:hit1}
\end{table}

\subsection{Ablation Studies}
\noindent\textbf{Impact of Embedding Feedback}: In PRASE, the PR module uses both the entity mappings and the embeddings as the feedback from the SE module. To explore the role of these two types of feedback, two additional PRASE frameworks are evaluated: one uses only the mapping feedback (denoted with suffix ``-M'') and the other uses only the embedding feedback (denoted with suffix ``-E'').
Table \ref{tab:vs} shows the comparison results. Generally, all three PRASE frameworks achieve good results, and the framework using both types of feedback performs the best on almost all datasets. 
Besides, the precision of the PRASE framework using only entity embeddings is higher than or close to the baseline PARIS, while the precision of the PRASE framework using only entity mappings is lower in most cases.
Furthermore, the use of entity mappings can significantly improve the recall but compromise the precision; the use of entity embeddings tends to maintain the precision, while the recall improvement is relatively limited. 
Table \ref{tab:vs} also shows that using both types of feedback can significantly improve the recall while maintaining the precision.

\noindent\textbf{Impact of Different SE Modules}: To analyze the impact of using different SE modules, four additional PRASE models are constructed based on MTransE, IPTransE, GCNAlign, and IMUSE. Table \ref{tab:PRASE_models} shows their performance. 
It can be observed that all the PRASE models achieve a higher F1-score than PARIS on all datasets except for PRASE-MTransE on MED-BBK-9K, which indicates that the PRASE framework is robust and not sensitive to the selection of the SE module. 
However, different SE modules may bring different degrees of improvement. Table \ref{tab:hit1} shows the Hits@1 of the entity mappings predicted by different SE modules on $\tilde{\mathcal{U}}^{\text{P}}$.
Their performance is positively correlated to the corresponding PRASE models. 
Specifically, BootEA and MultiKE outperform the other SE modules, and their corresponding PRASE models also perform better on most datasets.
Besides, the Hits@1 of MTransE is only 0.9\%, which is consistent with the degraded performance of PRASE-MTransE compared with PARIS.
We additionally perform PRASE-MTransE-E on MED-BBK-9K, and the F1-score reaches $0.511$, which shows that the entity embeddings can still provide useful information and benefit the PR module, even if the SE module has a poor alignment performance. Briefly, the benefit of the SE module is closely related to the selected embedding-based model. 
It is recommended to choose an advanced embedding-based model such as BootEA or MultiKE.
If the SE module has a poor alignment performance, it would be better to only use entity embeddings as the feedback to the PR module.

\begin{figure}[t!]
	\centering
	\includegraphics[width=1.0\linewidth]{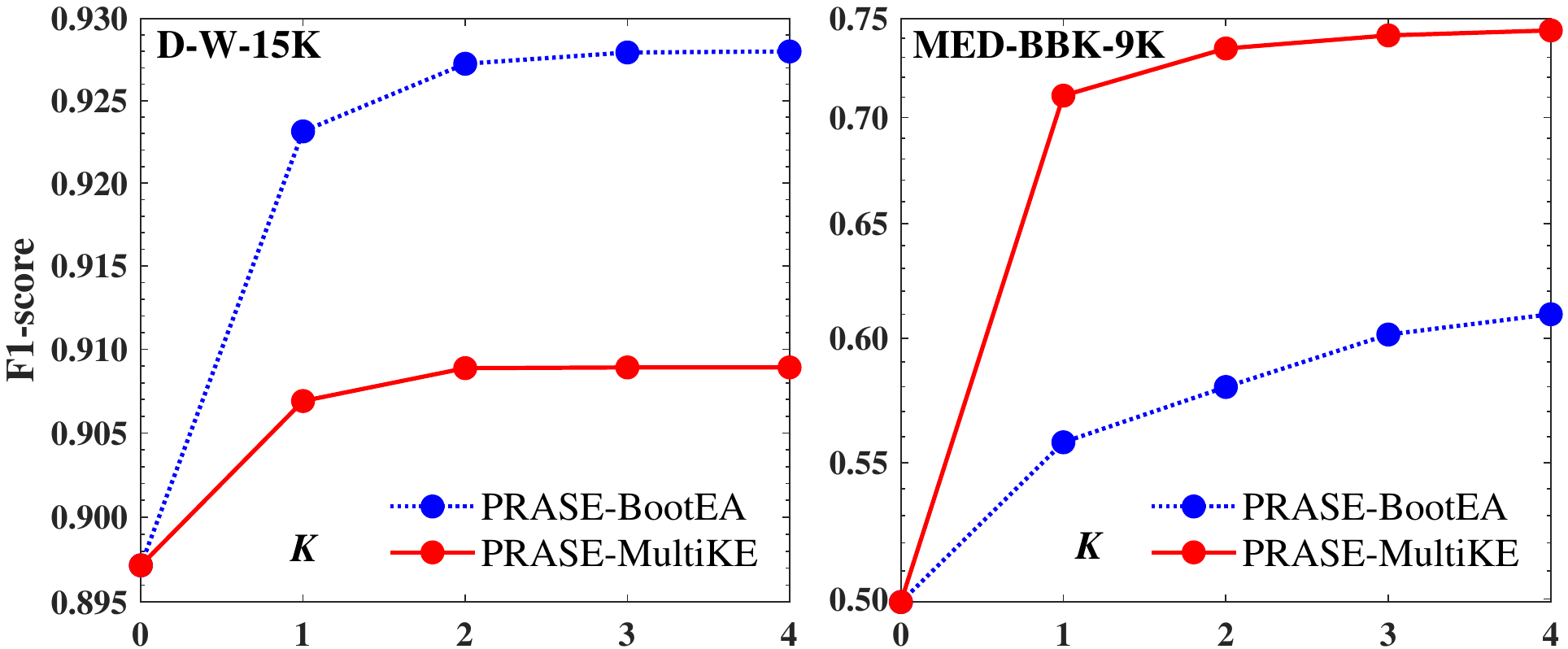}
	\caption{F1-score of PRASE models w.r.t. $K$ on two datasets.}
	\label{fig:Iteration}
\end{figure}

\noindent\textbf{Impact of Iteration Number}: To analyze the impact of the iteration number $K$, we additionally perform PRASE-BootEA and PRASE-MultiKE on D-W-15K and MED-BBK-9K with $K=2,3,$ and $4$. The results are shown in Figure \ref{fig:Iteration}, where the F1-score increases as $K$ increases. Therefore, more iterations may help to improve the PRASE model. However, in practice, since the framework converges quickly and the embedding learning in each iteration costs much more time, it is suggested to set $K$ to a small value.
\section{Conclusion and Discussion}
\label{conclusion}
In this work, an unsupervised KG alignment framework PRASE has been proposed, which consists of a probabilistic reasoning module, a semantic embedding module, and an iterative algorithm for the interaction of the two modules.
PRASE is compatible with most existing embedding-based models.
Extensive experiments on six datasets have verified the state-of-the-art performance of PRASE. More importantly, this work has shed light on the potential of unifying probabilistic reasoning and semantic embedding for KG alignment.
It is therefore necessary to call for such hybrids for academic research and industrial applications.
% As side results, PRASE outputs relation mappings through the relations' subsumption relationships by PR module. This will be further studied in the future. Meanwhile,
For future work, we plan to expand PRASE with other reasoning-based systems (e.g., LogMap) and enhance the interaction between the PR and SE modules by, e.g., injecting prior knowledge defined by the KGs' ontologies~\cite{chen2021augmenting}.
We will also utilize the alignment of KGs to address KG refinement problems such as error detection~\cite{chen2020correcting}.

% \section*{Acknowledgments}
% This work is supported by ...
\clearpage
\bibliographystyle{named}
\bibliography{ijcai21}

\begin{thebibliography}{}

\bibitem[\protect\citeauthoryear{Bordes \bgroup \em et al.\egroup
  }{2013}]{ANA2013NIPS}
Antoine Bordes, Nicolas Usunier, Alberto García-Durán, Jason Weston, and
  Oksana Yakhnenko.
\newblock Translating embeddings for modeling multi-relational data.
\newblock In {\em Proceedings of the 27th Annual Conference on Advances in
  Neural Information Processing Systems}, pages 2787--2795, Lake Tahoe, United
  States, December 2013.

\bibitem[\protect\citeauthoryear{Chen \bgroup \em et al.\egroup
  }{2017}]{MYM2017IJCAI}
Muhao Chen, Yingtao Tian, Mohan Yang, and Carlo Zaniolo.
\newblock Multilingual knowledge graph embeddings for cross-lingual knowledge
  alignment.
\newblock In {\em Proceedings of the 26th International Joint Conference on
  Artificial Intelligence}, pages 1511--1517, Melbourne, Australia, August
  2017.

\bibitem[\protect\citeauthoryear{Chen \bgroup \em et al.\egroup
  }{2020}]{chen2020correcting}
Jiaoyan Chen, Xi~Chen, Ian Horrocks, Erik B.~Myklebust, and Ernesto
  Jimenez-Ruiz.
\newblock Correcting knowledge base assertions.
\newblock In {\em Proceedings of The Web Conference 2020}, pages 1537--1547,
  April 2020.

\bibitem[\protect\citeauthoryear{Chen \bgroup \em et al.\egroup
  }{2021}]{chen2021augmenting}
Jiaoyan Chen, Ernesto Jimenez-Ruiz, Ian Horrocks, Denvar Antonyrajah, Ali
  Hadian, and Jaehun Lee.
\newblock Augmenting ontology alignment by semantic embedding and distant
  supervision.
\newblock In {\em Proceedings of the 18th Extended Semantic Web Conference},
  pages 1--16, Heraklion, Greece, June 2021.

\bibitem[\protect\citeauthoryear{Guo \bgroup \em et al.\egroup
  }{2019}]{LZW2019ICML}
Lingbing Guo, Zequn Sun, and Wei Hu.
\newblock Learning to exploit long-term relational dependencies in knowledge
  graphs.
\newblock In {\em Proceedings of the 36th International Conference on Machine
  Learning}, pages 2505--2514, Long Beach, United States, June 2019.

\bibitem[\protect\citeauthoryear{He \bgroup \em et al.\egroup
  }{2019}]{FZY2019DASFAA}
Fuzhen He, Zhixu Li, Yang Qiang, An~Liu, Guanfeng Liu, Pengpeng Zhao, Lei Zhao,
  Min Zhang, and Zhigang Chen.
\newblock Unsupervised entity alignment using attribute triples and relation
  triples.
\newblock In {\em Proceedings of the 24th International Conference on Database
  Systems for Advanced Applications}, pages 367--382, Chiang Mai, Thailand,
  April 2019.

\bibitem[\protect\citeauthoryear{Hogan \bgroup \em et al.\egroup
  }{2020}]{AEM2020ARXIV}
Aidan Hogan, Eva Blomqvist, Michael Cochez, Claudia d'Amato, Gerard de~Melo,
  Claudio Gutierrez, Jos{\'e} Emilio~Labra Gayo, Sabrina Kirrane, Sebastian
  Neumaier, Axel Polleres, Roberto Navigli, Axel-Cyrille~Ngonga Ngomo,
  Sabbir~M. Rashid, Anisa Rula, Lukas Schmelzeisen, Juan Sequeda, Steffen
  Staab, and Antoine Zimmermann.
\newblock Knowledge graphs.
\newblock {\em arXiv preprint arXiv:2003.02320}, 2020.

\bibitem[\protect\citeauthoryear{Jiménez-Ruiz and Grau}{2011}]{EB2011ISWC}
Ernesto Jiménez-Ruiz and Bernardo~Cuenca Grau.
\newblock {LogMap}: Logic-based and scalable ontology matching.
\newblock In {\em Proceedings of the 10th International Semantic Web
  Conference}, pages 273--288, Bonn, Germany, October 2011.

\bibitem[\protect\citeauthoryear{Kipf and Welling}{2017}]{TM2017ICLR}
Thomas~N. Kipf and Max Welling.
\newblock Semi-supervised classification with graph convolutional networks.
\newblock In {\em Proceedings of the 5th International Conference on Learning
  Representations}, pages 1--14, Toulon, France, April 2017.

\bibitem[\protect\citeauthoryear{Suchanek \bgroup \em et al.\egroup
  }{2012}]{FSP2012VLDB}
Fabian~M. Suchanek, Serge Abiteboul, and Pierre Senellart.
\newblock {PARIS}: Probabilistic alignment of relations, instances, and schema.
\newblock In {\em Proceedings of the 38th International Conference on Very
  Large Databases}, pages 157--168, Istanbul, Turkey, August 2012.

\bibitem[\protect\citeauthoryear{Sun \bgroup \em et al.\egroup
  }{2018}]{ZWQ2018IJCAI}
Zequn Sun, Wei Hu, Qingheng Zhang, and Yuzhong Qu.
\newblock Bootstrapping entity alignment with knowledge graph embedding.
\newblock In {\em Proceedings of the 27th International Joint Conference on
  Artificial Intelligence}, pages 4396--4402, Stockholm, Sweden, July 2018.

\bibitem[\protect\citeauthoryear{Sun \bgroup \em et al.\egroup
  }{2020}]{ZQW2020VLDB}
Zequn Sun, Qingheng Zhang, Wei Hu, Chengming Wang, Muhao Chen, Farahnaz Akrami,
  and Chengkai Li.
\newblock A benchmarking study of embedding-based entity alignment for
  knowledge graphs.
\newblock In {\em Proceedings of the 46th International Conference on Very
  Large Databases}, pages 2326--2340, Tokyo, Japen, August 2020.

\bibitem[\protect\citeauthoryear{Wang \bgroup \em et al.\egroup
  }{2017}]{wang2017knowledge}
Quan Wang, Zhendong Mao, Bin Wang, and Li~Guo.
\newblock Knowledge graph embedding: A survey of approaches and applications.
\newblock {\em IEEE Transactions on Knowledge and Data Engineering},
  29(12):2724--2743, 2017.

\bibitem[\protect\citeauthoryear{Wang \bgroup \em et al.\egroup
  }{2018}]{ZQX2018EMNLP}
Zhichun Wang, Qingsong Lv, Xiaohan Lan, and Yu~Zhang.
\newblock Cross-lingual knowledge graph alignment via graph convolutional
  networks.
\newblock In {\em Proceedings of the Conference on Empirical Methods in Natural
  Language Processing}, pages 349--357, Brussels, Belgium, October 2018.

\bibitem[\protect\citeauthoryear{Wu \bgroup \em et al.\egroup
  }{2019}]{YXY2019IJCAI}
Yuting Wu, Xiao Liu, Yansong Feng, Zheng Wang, Rui Yan, and Dongyan Zhao.
\newblock Relation-aware entity alignment for heterogeneous knowledge graphs.
\newblock In {\em Proceedings of the 28th International Joint Conference on
  Artificial Intelligence}, pages 5278--5284, Macao, China, August 2019.

\bibitem[\protect\citeauthoryear{Zhang \bgroup \em et al.\egroup
  }{2019}]{QZW2019IJCAI}
Qingheng Zhang, Zequn Sun, Wei Hu, Muhao Chen, Lingbing Guo, and Yuzhong Qu.
\newblock Multi-view knowledge graph embedding for entity alignment.
\newblock In {\em Proceedings of the 28th International Joint Conference on
  Artificial Intelligence}, pages 5429--5435, Macao, China, August 2019.

\bibitem[\protect\citeauthoryear{Zhang \bgroup \em et al.\egroup
  }{2020}]{ZJX2020COLING}
Ziheng Zhang, Jiaoyan Chen, Xi~Chen, Hualuo Liu, Yuejia Xiang, Bo~Liu, and
  Yefeng Zheng.
\newblock An industry evaluation of embedding-based entity alignment.
\newblock In {\em Proceedings of the 28th International Conference on
  Computational Linguistics}, pages 179--189, Barcelona, Spain, December 2020.

\bibitem[\protect\citeauthoryear{Zhu \bgroup \em et al.\egroup
  }{2017}]{HRZ2017IJCAI}
Hao Zhu, Ruobing Xie, Zhiyuan Liu, and Maosong Sun.
\newblock Iterative entity alignment via joint knowledge embeddings.
\newblock In {\em Proceedings of the 26th International Joint Conference on
  Artificial Intelligence}, pages 4258--4264, Melbourne, Australia, August
  2017.

\end{thebibliography}

\end{document}